\documentclass[conference]{IEEEtran}
\usepackage{header}
\usepackage{overpic}
\usepackage{amsmath}
\DeclareMathOperator{\NN}{NN}
\DeclareMathOperator{\ODESolve}{ODESolve}

\usepackage{multirow}

\begin{document}

\title{Physics-informed Machine Learning of Parameterized Fundamental Diagrams}

\author{\IEEEauthorblockN{James Koch\IEEEauthorrefmark{1},
Thomas Maxner\IEEEauthorrefmark{3},
Vinay Amatya\IEEEauthorrefmark{2}, 
Andisheh Ranjbari\IEEEauthorrefmark{4} and
Chase P. Dowling\IEEEauthorrefmark{2}}

\IEEEauthorblockA{Pacific Northwest National Laboratory\\
          \IEEEauthorrefmark{1}National Security Directorate, Seattle, WA, USA \\ }
          
\IEEEauthorblockA{\IEEEauthorrefmark{2}Physical and Computational Sciences Directorate, Seattle, WA, USA \\
          \{\tt{james.koch}\},\{\tt{vinay.amatya}\},\{\tt{chase.dowling}\}\tt{@pnnl.gov} }
          
\IEEEauthorblockA{\IEEEauthorrefmark{3}University of Washington\\
Department of Civil and Environmental Engineering, 
Seattle, WA, USA \\
\tt{tmaxner@uw.edu}}

\IEEEauthorblockA{\IEEEauthorrefmark{4}Pennsylvania State University\\
Department of Civil and Environmental Engineering, 
State College, PA, USA \\
\tt{ranjbari@psu.edu}}
}


\maketitle

\begin{abstract}

Fundamental diagrams describe the relationship between speed, flow, and density for some roadway (or set of roadway) configuration(s). These diagrams typically do not reflect, however, information on how speed-flow relationships change as a function of exogenous variables such as curb configuration, weather or other exogenous, \emph{contextual} information. 
In this paper we present a machine learning methodology that respects known engineering constraints and physical laws of roadway flux--those that are captured in fundamental diagrams--and show how this can be used to introduce contextual information into the generation of these diagrams. The modeling task is formulated as a probe vehicle trajectory reconstruction problem with \textit{Neural Ordinary Differential Equations} (Neural ODEs).
With the presented methodology, we extend the fundamental diagram to non-idealized roadway segments with potentially obstructed traffic data. For simulated data, we generalize this relationship by introducing contextual information at the learning stage, i.e. vehicle composition, driver behavior, curb zoning configuration, etc, and show how the speed-flow relationship changes as a function of these exogenous factors independent of roadway design. 

\end{abstract}

\section{Introduction}

The fundamental diagram measures the relationships between average vehicle speed, density, and flux through a boundary for a given roadway in an idealized environment \cite{greenshields1935study}. Nearing a century of study and use, these diagrams formed the foundation for road design and selection, while more recently they enable optimization routines in vehicles employing adaptive cruise control on roadways with connected sensor and communications networks \cite{Talebpour2016}. Given rapid expansion in available roadway and vehicular data sources, we propose that fundamental diagrams can be learned that are specific to special cases of roadways that are otherwise superficially similar. For example, a city avenue and airport departures drive may have a similar number of lanes, lane widths, and speed limit, but drivers experience very different patterns of vehicular flow due to turbulent curb activity on an airport departures drive and differences in average vehicle class composition. Essentially, the driving difference between these roadways is the curb use and configuration, but this \textit{contextual} information is generally not reflected in fundamental diagrams. Features like weather, grade, network topology, vehicle composition, and driver behavior potentially alter the fundamental diagram of a section of roadway. In this paper we propose a machine learning methodology that 1) assumes a functional structure commensurate with known physical relationships governing the flow of vehicles of roadways and 2) combines vehicle trajectory and static sensor data to learn a fundamental diagram for a roadway segment. We show how that methodology can then be used to compare and accept in a principled fashion as input features or additional output dimensions, contextual information such as curb configuration. Thus, the specific contributions of this work are (i) recasting the regression of vehicle data to various fundamental diagrams as a trajectory reconstruction problem, (ii) the generalization of fundamental diagrams to exogenous variables, and (iii) the methodology for accomplishing these modeling tasks. 

A brief review of fundamental diagrams and their application is presented in Section \ref{sec:background}. Following this, the methodology for this study is given in Section \ref{sec:methods} and associated results in Section \ref{sec:results}. Lastly we discuss results and methodology in Section \ref{sec:discussion}.

\begin{figure*}[]
        \centering
            \begin{overpic}[width=\linewidth]{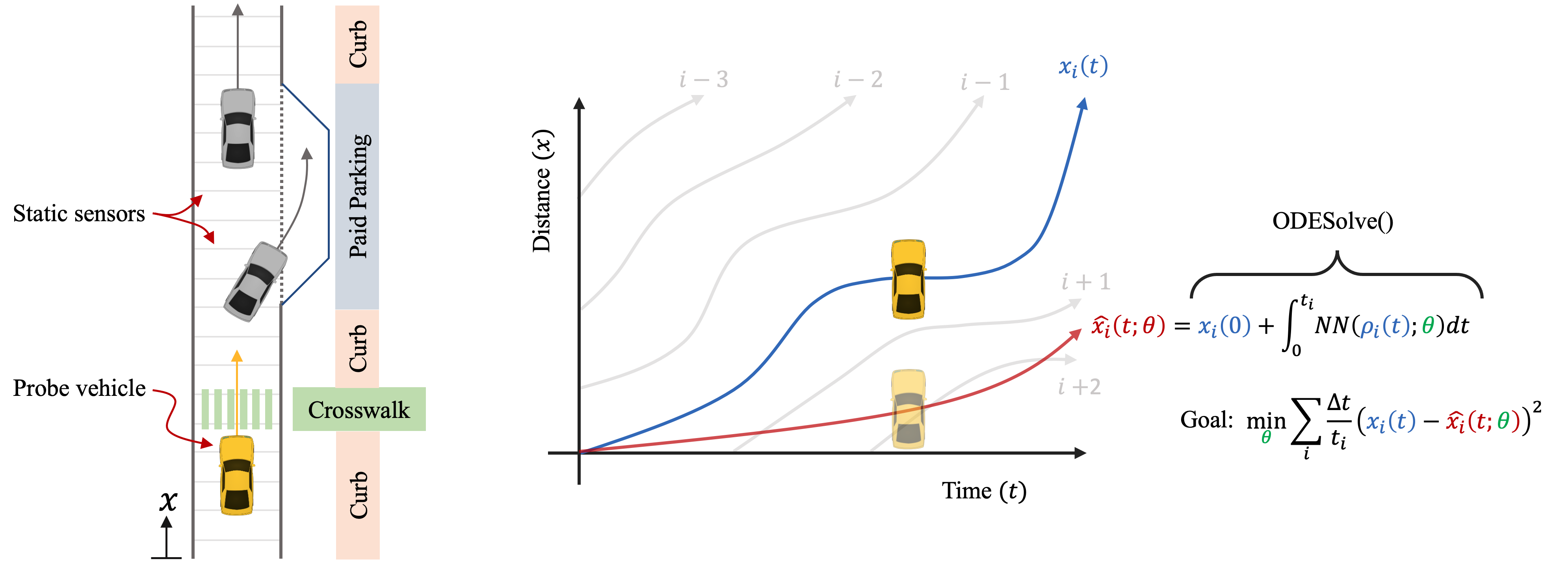}  
            \put(5,35){(a)}
            \put(33,35){(b)}
	    \end{overpic}  
	    \caption{The layout of a typical urban street may not differ substantially from block-to-block. However, the interface of these streets with the curb varies drastically among a population of otherwise similar streets. Driver behavior on these streets can vary in an equally dramatic manner, suggesting the need for a local, parameterized extension of the \textit{Fundamental Diagram} to account for these exogenous inputs. In (a), shown is a notional street and curb configuration whose interaction contributes to driver behavior (e.g. availability of parking or the presence of a crosswalk). A high-level overview of our method for learning parameterized relationships between speed, density, and a particular roadway is shown in (b). The regression task is formulated as a trajectory reconstruction problem, where the evolution of a probe vehicle's position is governed by the transition dynamics specified by the learnable fundamental diagram ($\NN(\rho,\ldots)$). In this pictorial representation of the learning process, the probe vehicle's trajectory $x_i \left( t \right)$ is approximated by $\hat{x_i} \left( t; \theta \right)$. We seek to minimize the difference between these two trajectories in $x-t$ space. }
		\label{fig:intro}
\end{figure*}

\section{Background} \label{sec:background}

Nearly a century of research on traffic flow theory has been conduced since Greenshield's original formulation relating traffic flow quantities of average vehicle speed, density, and vehicle flux \cite{greenshields1935study, turner201175} on a segment of roadway; developments since that time have focused along two key dimensions: spatial coverage and data source. 

\begin{figure}
    \centering
    \includegraphics[width=\columnwidth]{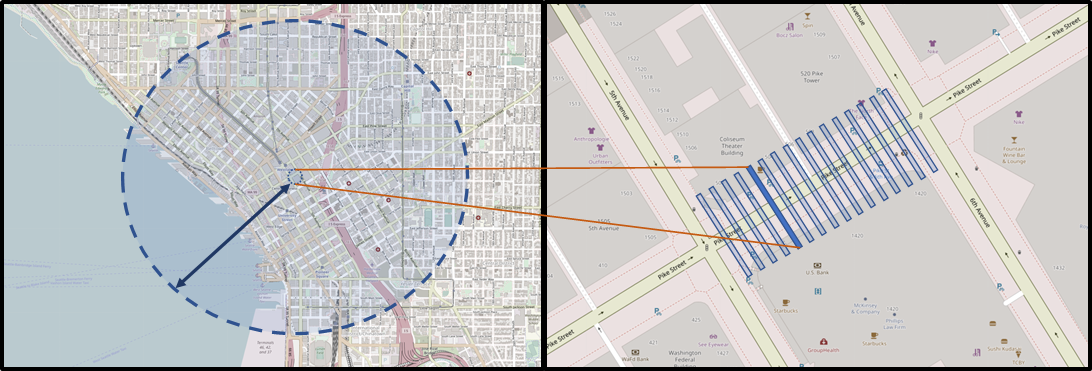}
    \caption{The conceptual distinction between macroscopic and microscopic fundamental diagrams: the size of the boundary through which flux is computed. Macroscopic FD's draw a large boundary represented by the dashed circle, averaging speed-flow information within the boundary but requiring less high resolution data to make conclusions about constituent roadway segments (e.g. sparse vehicle probe data traversing the closed region). Microscopic FD's on the right trade off intense data requirements for greater resolution on the speed-flow relationships unique to a specific segment of roadway. Several boundaries can be averaged in sequence to produce a link fundamental diagram, similar in spirit to drawing a closed boundary around the link, but the key conceptual difference is the size of the boundary through which flux is computed.}
    \label{fig:macromicro}
\end{figure}

Historically, traffic flow data on a roadway has been expensive to acquire in contrast to today: significant attention has thus been paid to \emph{efficiently} measuring traffic flow relationships. Addressed from a spatial perspective, fundamental diagrams were computed with very high fidelity for select few, frequently representative, locations such as a segment of a straight highway stretch, necessitating only a single concentrated data collection effort to model all similar sections of roadway, as reflected for example in the Highway Capacity Manual \cite{HCM2000}. Alternatively, fundamental diagrams have been computed with data averaged over a larger area: so-called macroscopic fundamental diagrams \cite{daganzo2007urban, daganzo2008analytical}, such that an average traffic relationship could be estimated with sparse samples across a large number of roadways segments within a closed boundary. The difference in spatial dimension is best conceptualized in Fig.~\ref{fig:macromicro}---in essence, a closed boundary through which (implicitly or explicitly) relevant traffic flow quantities of vehicle average speed, density, and flux are measured determines the relative resolution of the fundamental diagram learned describing that location's traffic flow.

Data collected on traffic flow quantities at these locations falls into two categories: vehicular data and static sensor data \cite{turner201175}. The difference in sources amounts to differences in reference frame by which analyses are conducted. Fundamental diagrams computed from vehicular data, often in the form of some probe vehicle, leverage a kinematic relationship between surrounding vehicles; while static sensors like cameras or tube counters, measure a fixed, site-specific point estimate of traffic flow quantities.

With these two dimensions---spatial coverage and data source---a large portion of works on modeling traffic flow theory can be categorized. For large spatial areas and sparse trajectories of vehicular probe data, so-called macroscopic fundamental diagrams use vehicle trajectory data to compute an aggregate fundamental diagram over a large area \cite{daganzo2007urban, daganzo2008analytical}, while recent works disaggregate and approximate individual fundamental diagrams on constituent roadway segments within a similarly large area \cite{geroliminis2011properties, maass2020street}. For localized segments of roadway, some recent works have defined much smaller closed boundaries to the same end, for example so-called \emph{link} fundamental diagrams \cite{Yu2020}. 

Combining both vehicular probe data and static sensor data---hybridizing reference frames--- \cite{moutari2007hybrid, claudel2008guaranteed, leclercq2007hybridization}. This approach frequently appears in works aimed at connected autonomous vehicles executing a driving control law that seeks to improve traffic flow at the link level, as modeled by fundamental diagrams of speed and flow \cite{wu2018stabilizing, kreidieh2018dissipating}. Both the reference frame of the individual probe vehicle collecting local data and relatively global information in the neighborhood of that vehicle dervied from static sensors or other connected vehicles on a roadway link need to be considered simultaneously to enact a control law, where the impact of the law on traffic flow is a function of a segment-specific fundamental diagram. 

The missing category in existing traffic flow modeling works involves combining both vehicular and static sensor data over a large spatial area due to the intense data costs involved, with the spatial granularity of static sensors being the binding constraint. To achieve high spatial resolution over a large geographic area, however, requires a general method for computing fundamental diagrams that can accept as input features contextual data specific to each location. Thus, we propose a general method for modeling traffic flow amenable to any desired spatial resolution and available data type, vehicular, static sensor, or both. In this paper we will focus on variations of a specific roadway segment to demonstrate that the method is agnostic to specifics and can learn contextual features while taking the powerful hybridizing approach of combining both vehicle probe and static sensor data. Generalizing from the link fundamental diagram of \cite{wu2018stabilizing}, we introduce a contextual fundamental diagram that, while assuming Greensheilds' relationship or more general physical models of traffic flow \emph{a priori}, can include amongst empirical traffic flow measurements, arbitrary contextual information such as weather, road grade, vehicle composition, and so on, that might impact the traffic flow characteristics of a roadway segment. Thus, microscopic fundamental diagrams can be learned that are specific to unique roadway segments in a fashion specified entirely by the type and resolution of data provided.

\begin{figure}
    \centering
    \includegraphics[width=\columnwidth]{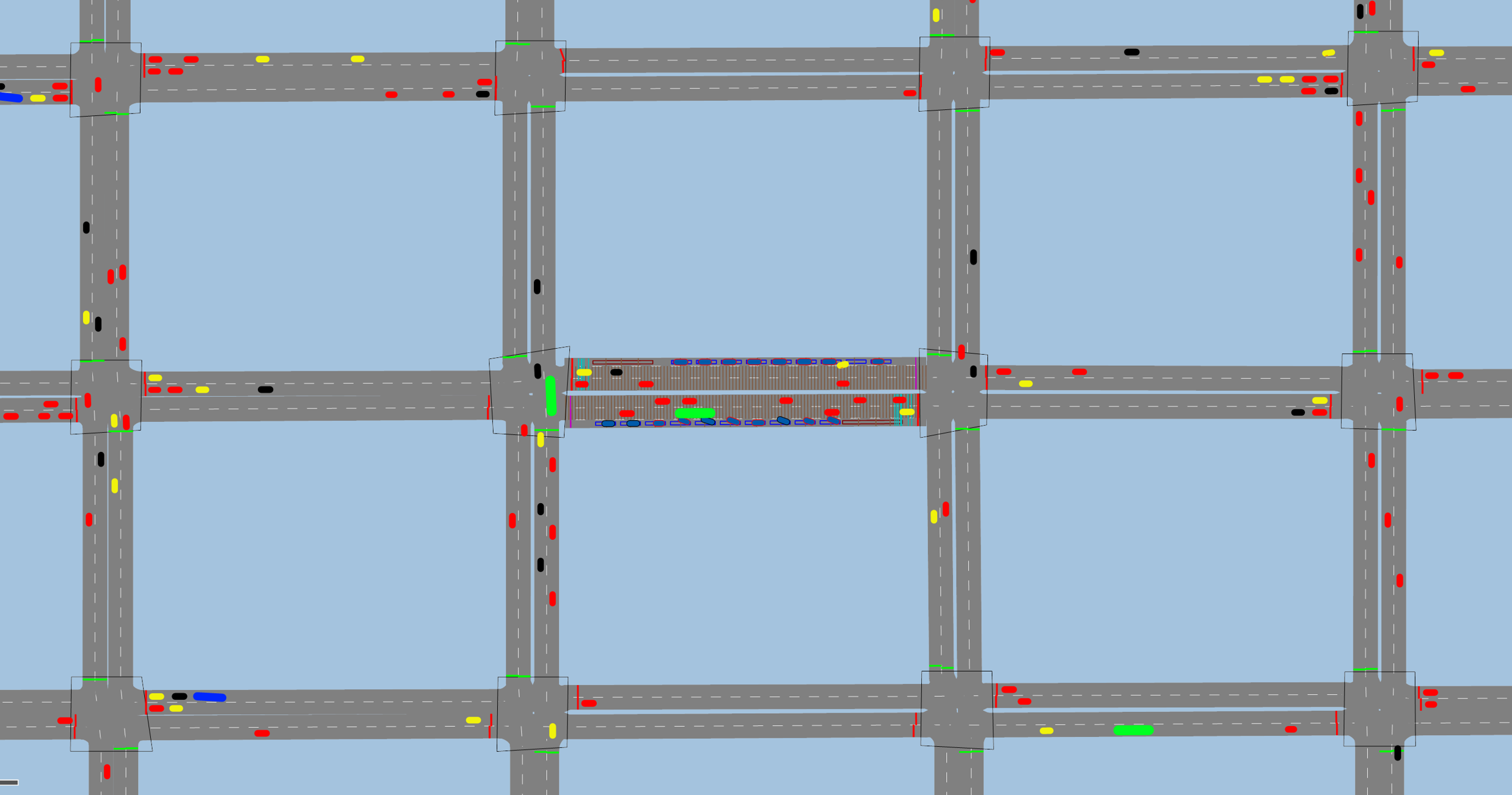}
    \caption{Illustration of homogeneous city-block network used in our VISSIM simulations; centered street segment is the segment over which a microscopic fundamental diagram is computed.}
    \label{fig:simulator}
\end{figure}

\subsection{Fundamental Diagram Preliminaries}

Appealing to the fluid dynamics analogy of Lighthill and Whitham \cite{lighthill1955kinematic}, the bulk motion of vehicles in one spatial dimension can be expressed as the conservation of vehicles, a partial differential equation (PDE) evolving density $\rho(x,t)$ (with units of vehicles per unit length) through space $x$ and time $t$:

\begin{equation} \label{eq:conservation}
    \frac{\partial \rho}{\partial t} + \frac{\partial Q }{\partial x} = 0,
\end{equation}
where the flux $Q=\rho u$. The relationship between local density and velocity is:
\begin{equation} \label{eq:FD}
    u = f\left( \rho; \theta \right):\mathbb{R}^{1+n} \rightarrow \mathbb{R}^1,
\end{equation}
where $f\left( \cdot \right)$ is the \textit{Fundamental Diagram} (FD) of traffic flow for a particular considered situation parameterized by the length-$n$ parameter vector $\theta$.

The solution of this conservation equation is a space-time field of density. A particle's path (i.e. a vehicle in this setting) can be found by solving the differential equation:

\begin{equation} \label{eq:particle_path}
    \frac{dx\left(t\right)}{dt} = f\left(\rho\left(x\left(t\right),t\right);\theta\right).
\end{equation}
Succinctly, the instantaneous velocity of a vehicle is given by the FD and the experienced density along the path $x\left(t\right)$ of the vehicle.

The specific shape of the FD (e.g. a concave-down flux-speed relationship) can give rise to many distinct qualitative behaviors of traffic flow. These include shock and rarefaction waves - i.e. canonical ``stop-and-go'' traffic jams or jamitons \cite{seibold2012constructing} - that propogate as a wave at speeds derived from the characteristics of the PDE of Eq. \ref{eq:conservation}.

The simplest FD that faithfully reproduces the qualitative behavior seen on real roadways is that of Greenshield. Greenshield's model is a simple linear relationship between speed and density:
\begin{equation} \label{eq:green}
    f_G \left(\rho \right) = 
     \begin{cases} 
      u_0 \left( 1 - \frac{\rho \left(t\right)}{\rho_j} \right) & \rho < \rho_j \\
      0 &  \rho \geq \rho_j 
   \end{cases}
\end{equation}
where $u_0$ is freeflow speed and $\rho_j$ is the jam density of the flow. The Greenshield FD is attractive because of its simplicity and ease of parameter selection. However, with its simplicity comes reduced expressivity. For example, by construction, the linear speed-density relationship gives a capacity density equal to half of the jam density; this is typically an over-estimation \cite{turner201175}. 

Piecewise FDs increase model flexibility by allowing for different modes of traffic flow. For example, in the Triangle FD \cite{greenberg1959analysis}, a free-flow speed and density range are defined to provide better low-density model performance over that of Greenshield's FD. Similarly, discontinuous FDs aim to introduce sharp boundaries between modes of traffic flow to account for different qualitative behaviors \cite{edie1961car,wong2002multi}, e.g. during free-flow and during queuing discharge. This constitutes the distinction between single-regime and multi-regime diagrams.

\subsection{Model Construction and Fitting}

Many methods exist for fitting, calibrating, and fine-tuning the wide array of FDs in literature. The vast majority focus on regressing a chosen FD functional form directly to point cloud data (i.e. curve fitting) in the speed-density or flux-density coordinate systems, depending on the available data. Fitting Greenshield's FD (or any other single-regime FD prior) amounts to performing least squares regression to speed-density data pairs (e.g., as in \cite{QU201591}). Other methods for fitting these data pairs include Maximum Likelihood Estimation \cite{9703273} (MLE) and Mixed-Integer Programming \cite{li2011fundamental} (MIP).

\subsection{Machine Learning in Traffic Modeling}
Departing from the standard curve-fitting task, novel Machine Learning (ML) approaches leverage autodifferentiation to fit models to available data. The advantages of using ML in this context are (i) the increased flexibility of universal function approximators to handle data from a variety of sensors, resolutions, and formats, (ii) the ability to encode physics as model priors, and (iii) the ease of model construction. Here, ML methods generally fall into one of two categories; \textit{black-box} or \textit{physics-informed}. In the former, models are constructed and trained without regard for satisfying physical priors or other constraints. Their objectives are solely to minimize prediction loss. As an example of a black-box ML model, in \cite{9740401} a Convolutional Neural Network (CNN) with novel kernel designs is trained to ingest vehicle trajectories and output reconstructed $x-t$ \textit{continuous} velocity fields. Physics-informed ML models aim to incorporate known physics in the construction and/or training of a model. For example, given the known physics of Equations \ref{eq:conservation}-\ref{eq:FD}, one can construct a \textit{Physics-Informed Neural Network} (PINNs \cite{RAISSI2019686}) to reconstruct the density field given appropriate measurements. In \cite{9531557},  point measurements of density in $x-t$ space are used to automatically reconstruct the scalar density field and provide an estimate of the FD using a PINN that is trained to minimize the residual of the assumed governing equations. Similarly, in \cite{9565096}, the authors perform traffic state estimation using measurements from probe vehicles in addition to static sensors. In this work, the authors choose a FD prior to fit to the available data subject to a PDE (Eq. \ref{eq:conservation}) residual.

Physics-informed ML-based methods in this context primarily rely upon the reconstruction and residual of the governing PDE to guide the optimization to a high-quality model. In many situations, this is difficult because of the sparsity of measurements in the domain of interest, leading to massively ill-posedness of the inverse problem. Thus, these methods depend on (i) the amount of data, (ii) their expanse on the domain of interest, and (iii) its quality (noise).

In this work, we depart from the standard curve-fitting task as well as from the PDE-based physics-informed ML paradigm. Instead, we leverage vehicle trajectory data following the assumed vehicle kinematics of Eq. 
\ref{eq:particle_path}. Provided that speed-density pairs can be provided along a vehicle's trajectory through $x-t$ space, right-hand-side of Eq. \ref{eq:particle_path} (a FD specification) can be inferred directly by using \textit{Neural Ordinary Differential Equations}, or Neural ODEs \cite{chen2018neural}. Exploiting vehicle kinematics in this manner has several benefits. First, a FD can be estimated from as little as a single vehicle trajectory and the learning method is easily extensible to an arbitrary number of probe vehicles. Second, there is no need to perform a reconstruction of the entire $x-t$ density and velocity spatial fields. Third, a trajectory reconstruction task (as opposed to a standard curve fit) carries with it the constraint of kinematic consistency: a least-squares curve fit of a FD does not necessarily produce the FD that minimizes the trajectory reconstruction error. 

\section{Methods} \label{sec:methods}

\subsection{Neural Ordinary Differential Equations}
Our modeling goal is to regress vehicle trajectory and loop detector data to FDs of various complexity by minimizing the trajectory reconstruction error. Neural ODEs are used for this task. Neural ODEs evolve a system's state $\mathbf{z}$ through an independent variable (typically time $t$) via some dynamics specified by a learnable function approximator or composite function \cite{rackauckas2020universal}. A neural ODE is specified as:
\begin{equation} \label{eq:node}
\frac{d \mathbf{z}}{dt} = f\left( \mathbf{z}, t; \theta \right), 
\end{equation}
where $f$ is the input-output mapping of the system state to be learned and $\theta$ are the trainable parameters. Note that as written, Eq. \ref{eq:node} is a non-autonomous ODE; that is, there is an explicit dependence on the independent variable (such as through forcing or a control input).  The Initial Value Problem (IVP) corresponding to Eq. \ref{eq:node} can be solved by integrating:
\begin{equation}
    \mathbf{z}\left(t_{end}\right) = \mathbf{z}\left(t_0\right) + \int_{t_0}^{t_{end}} f\left( \mathbf{z}(t), t; \theta \right) dt.
\end{equation}
This IVP can be solved numerically with standard ODE solvers:
\begin{equation}
    \mathbf{z}\left(t_{end}\right) = \ODESolve(f,\mathbf{z}(t_0),t_0, t_{end};\theta).
\end{equation}
The parameter vector $\theta$ is updated by obtaining gradients of the a residual with respect to the parameters and performing gradient descent. This is achieved through (i) composing a computational graph of elementary operations of the ODE solver (e.g. 4th order Runge Kutta) and performing backpropagation, or (ii) via the adjoint sensitivity method, whereby a second ODE for the residual is defined and solved backward in time \cite{chen2018neural}. 

For the trajectory reconstruction problem, the system state is probe vehicle location; $\hat{x}(t)$, where $\hat{\left(\cdot\right)}$ denotes model quantities (to distinguish them from those associated with the data). Furthermore, each vehicle trajectory has a corresponding density control input $\rho_c\left(t\right) = \rho \left(x\left(t\right),t\right)$, that is, the control input to the Neural ODE is the time series of the density experienced by a probe vehicle along its path $x(t)$ through the scalar field $\rho\left(x,t\right)$. Thus, we construct and train nonautonomous models of the form:
\begin{equation} \label{eq:node_fd}
    \frac{d \hat{x}\left(t\right)}{dt} = f\left( \hat{x}, \rho_c\left(t\right) ;\theta\right).
\end{equation}

\subsection{Simulation, Data, and Pre-Processing}

Data is collected through the PTV Vissim microscale traffic simulation platform. Vissim is a particle simulator where vehicle location, velocity, and acceleration, with respect to an individual driver's modeled behavior, is iteratively computed at a high time resolution. Figure \ref{fig:simulator} illustrates the test network and roadway segment (center) where data is collected. Static sensors are placed every 1 meter along the length of the roadway, which measures the presence of a vehicle and the vehicles entry and exit times for that sensor site. Additionally, vehicle trajectory data is recorded on vehicle position, velocity, and acceleration at a 1 second resolution. Simulations are generated for total input vehicle volumes ranging between 0 vehicles per hour and 600 vehicles per hour at each entry point, vehicle speed distributions between 20 and 30 kilometers per hour, and parking demand rates between 0 and 50\% of total vehicles present in the network. An exhaustive detailing of simulation parameters and copies of VISSIM configuration files used for independent verification and study replication are available in our code and documentation at \url{https://github.com/pnnl/neuralFD}.

\begin{figure}[]
        \centering
            \begin{overpic}[width=\columnwidth]{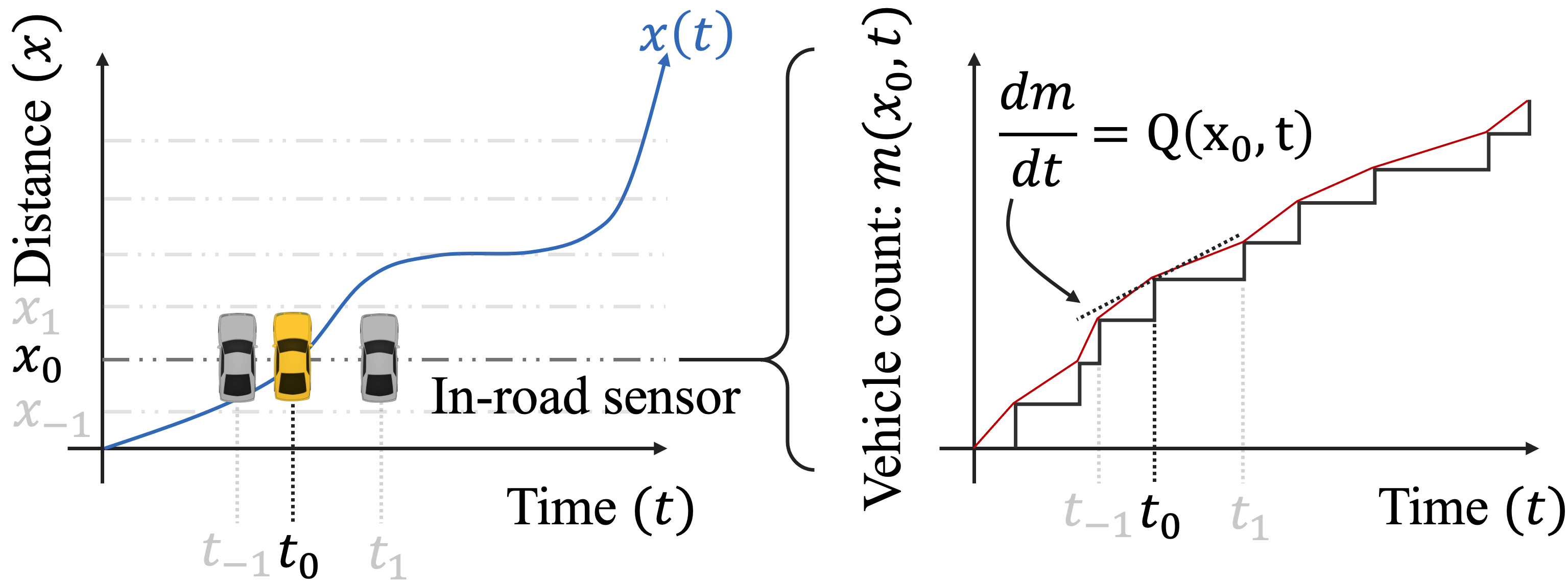}  
            \put(4,38){(a)}
            \put(60,38){(b)}
	    \end{overpic}  
	    \caption{In (a), depicted is an array of equally-spaced in-road sensors; i.e. loop detectors. The data from these sensors are event-driven. As a vehicle passes the sensor, the timestamp is recorded. Defining the unit of mass on a roadway as a vehicle, flux is therefore vehicles per unit time. Thus, an approximate measure of flux through time can be obtained (shown in (b)) by differentiating the (linearly interpolated) event data from the in-road sensors.}
		\label{fig:count}
\end{figure}

For our regression task, each simulation contains several hundred to thousands of unique vehicle trajectories. For each trajectory, a time series of co-located location and density data points is required. Figure \ref{fig:count} depicts the procedure for extracting flux time series from the in-road static sensors. Each sensor is event-driven; i.e. time stamps are recorded for every vehicle entry and exit. An approximate measure of flux through time can be obtained by differentiating the event data (linearly interpolated in space and time). An estimate for vehicle-experienced density is therefore $\rho(x,t) = Q(x,t)/u(x,t)$. The dataset consists of a collection of vehicle trajectories, each containing co-located space-density data pairs. For this study, from each scenario we randomly sample 500 vehicle trajectories to be used for training and randomly sample 100 of the remaining vehicle trajectories as a test set. 

\subsection{Model Formulations}
To demonstrate our thesis, we construct three separate FDs using the Neural ODE modeling paradigm. The first is the two-parameter Greenshield diagram, followed by one-variable and parameterized two-variable neural network-based FDs. Their formulations are as follows:

\subsubsection*{Greenshield's FD} The Greenshield FD takes two parameters to fully define the velocity-density relationship; a free-flow speed $u_0$ and a jam density $\rho_j$. The neural ODE takes the form:

\begin{equation} \label{eq:green_node}
    \frac{d \hat{x} \left( t \right)}{dt} = 
     \begin{cases} 
      u_0 \left( 1 - \frac{\rho_c \left(t\right)}{\rho_j} \right) & \rho_c < \rho_j \\
      0 &  \rho_c \geq \rho_j 
   \end{cases}
\end{equation}

\subsubsection*{Neural Network-Based FDs} We construct Deep Neural Networks (NNs) to use as trainable universal function approximators. For this study, networks $\NN_\theta \left( \mathbf{z} \right) : \mathbb{R}^m \rightarrow \mathbb{R}^1$ maps the $m$-dimensional input $z \in \mathbb{R}^m$ to a scalar output. The network is parameterized by the weights and biases of the successive layers; $\theta = \{ \mathbf{W}_0, \mathbf{b}_0, \dots ,\mathbf{W}_K, \mathbf{b}_K \}$ for layers $1 \leq k \leq K$:

\begin{align}
    \NN_\theta \left( \mathbf{z} \right) &= g \circ f_K \circ \dots \circ f_1 \left( \mathbf{z} \right) \\
    f_i &: \mathbb{R}^{n_i} \rightarrow \mathbb{R}^{n-1} \\
    f_i \left(\mathbf{z} \right) &= \sigma\left(\mathbf{W}_i \mathbf{z} + \mathbf{b}_i \right)
\end{align}
where $\mathbf{\sigma} : \mathbb{R} \rightarrow \mathbb{R}$ is the element-by-element application of an activation function $\sigma$ (e.g. hyperbolic tangent, rectified linear unit, sigmoid, etc.) and $g$ is the output layer function.

For the present work, we separately construct and train a single-variable neural network $\NN_{\theta}^1 \left(\frac{\rho}{\rho_j}\right)$ and a two-variable neural network $\NN_{\theta}^2 \left(\frac{\rho}{\rho_j}, x\right)$. The inputs are scaled according to a user-specified $\rho_j$ (0.05 veh./ft. for this study) such that the inputs and outputs are of the same order of magnitude. Each network is composed of two hidden layers of 10 nodes each. All activation functions are sigmoidal, including that of the output layer. Thus, $\NN_\theta^1 : \mathbb{R}^1 \rightarrow (0,1)$ and $\NN_\theta^2 : \mathbb{R}^2 \rightarrow (0,1)$.

The single-variable neural network-based neural ODE is therefore:
\begin{equation} \label{eq:single}
    \frac{d \hat{x} \left( t \right)}{dt} = u_0 \cdot \NN_{\theta}^1\left(\frac{\rho_c\left(t\right)}{\rho_j}\right) ,
\end{equation}
where $u_0$ is a trainable parameter corresponding to free-flow speed. In this manner, the output of the neural network is rescaled to $(0,u_0)$. Likewise, the two-variable neural ODE is:
\begin{equation} \label{eq:param}
    \frac{d \hat{x} \left( t \right)}{dt} = u_0 \cdot \NN_{\theta}^2\left(\frac{\rho_c\left(t\right)}{\rho_j},\hat{x} \right),
\end{equation}
which contains the same rescaling by $u_0$. 

Note that in this study, we have not varied the neural network topology (number of layers and nodes) nor have we performed hyperparameter selection tasks (activation functions, learning rate, etc.) Neural ODEs.

\subsection{Regression} \label{sec:training}
The regression tasks aim to minimize the residual between the reference trajectories, $\mathbf{x}_i \left(t\right)$, and the predicted trajectories, $\hat{\mathbf{x}_i} \left(t ; \theta \right)$:
\begin{equation} \label{eq:loss}
    \mathcal{L}\left(\theta\right) = \frac{1}{N} \sum_i^N \frac{\Delta t}{t_i} \left(\mathbf{x}_i - \hat{\mathbf{x}_i}\left(\theta\right) \right)^2 ,
\end{equation}
where $N$ is the number of trajectories in the training set or test set, $\Delta t$ is the temporal spacing between data points, and $t_i$ is the transit time associated with the $i$-th trajectory. For uniform temporal resolution, this cost function reduces to Mean Squared Error (MSE) on a per-trajectory basis (units of length squared).

The ``forward pass'' of the model generates a trajectory $\hat{\mathbf{x}_i} (\theta)$ via the numerical integration of Eq. \ref{eq:node_fd} evaluated at the co-location points of the reference trajectory's time series:
\begin{equation}
        \hat{\mathbf{x}}\left( \theta \right) = \ODESolve(f,\rho(t),\mathbf{x}(t_0),t_0, t_{end};\theta) \rvert_{t \in T} .
\end{equation}
Because each trajectory may have a different time span, the forward pass of the model involves an $\ODESolve()$ call for each trajectory in the data split. For each numerical experiment, the Adam optimizer with a learning rate of 0.1 was used until converged (small relative epoch-to-epoch change). After each epoch, the model was evaluated against the test split; the weights associated with the lowest test loss were selected for reporting.

\section{Results} \label{sec:results}
\begin{figure*}[]
        \centering
            \begin{overpic}[width=5in]{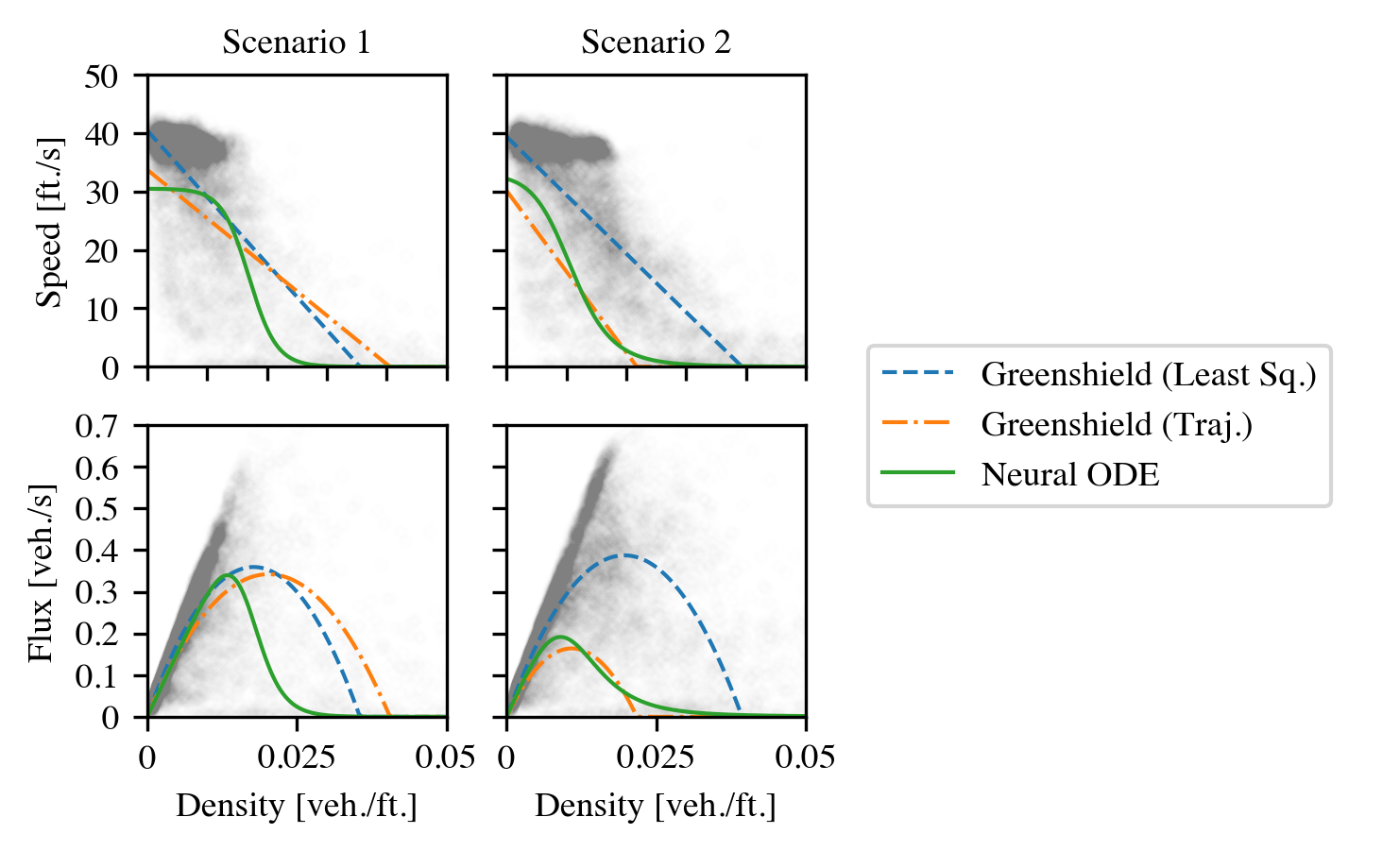}  
            \put(28,52){(a)}
            \put(53.5,52){(b)}
            \put(28,27){(c)}
            \put(53.5,27){(d)}
	    \end{overpic}  
	    \caption{Displayed are three Fundamental Diagrams tuned to match the underlaid speed-density data pairs for Scenario 1 (compliant drivers) and Scenario 2 (high rate of non-compliance and increased parking demand). Greenshield's FD was fit to the data using standard least-squares regression as a baseline model for comparison. A Neural ODE-formulation of Greenshield's diagram was tuned to minimize the trajectory reconstruction error of the probe vehicles. Lastly, a neural network-based Neural ODE (i.e. no imposed model structure other than kinematic consistency) was trained on the same data with the same trajectory minimization objective. Table \ref{tab:metrics} summarizes the performance of these three models. While the least-squares minimization problem captures the scatter in the raw data pairs, it is least performant in the trajectory reconstruction task.}
		\label{fig:single_variable}
\end{figure*}
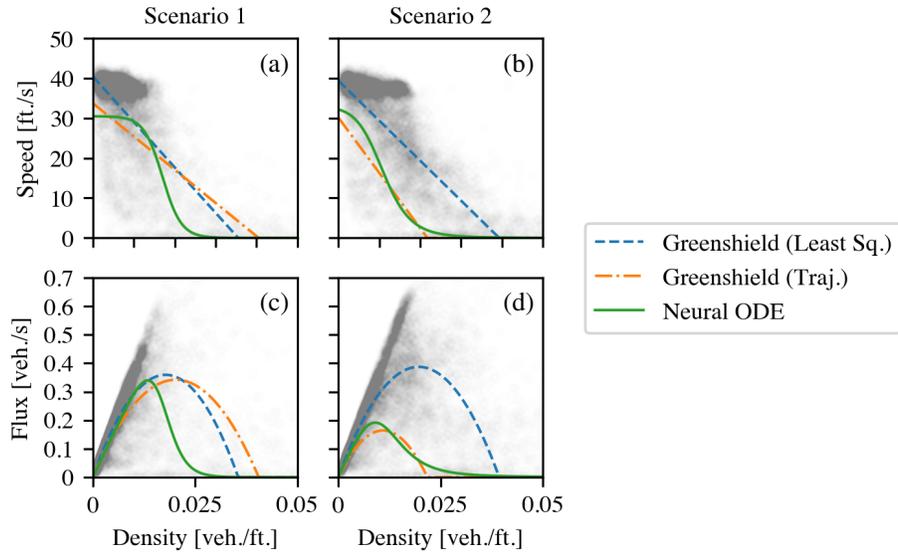

\begin{table*}

\caption{Model selection metrics. Boldface represents the best (lowest) value among the candidate models. Error units are square feet.}

\label{tab:metrics}
\centering
\begin{tabular}{lc|c|c|c|c}
                                                 &  & Greenshields (Least Sq.)    &  Greenshields (Traj.) & $\NN(\rho) $ & $\NN(\rho,x) $ \\ \hline
\multicolumn{1}{l|}{\multirow{2}{*}{Scenario 1}} & Train & 6464 &     5069       &      3836      &        \textbf{562.0}       \\
\multicolumn{1}{l|}{}                            & Test  & 3437 &     3029       &      2100      &        \textbf{717.0}       \\ \hline
\multicolumn{1}{l|}{\multirow{2}{*}{Scenario 2}} & Train & 56600 &    17210        &     16560   &          \textbf{1445}     \\
\multicolumn{1}{l|}{}                            & Test  & 106250 &      13080      &      11250      &         \textbf{2015}     
\end{tabular}
\end{table*}

\subsection{Simulating Speed-Flow Impacts due to Curb Congestion}

As a use-case for a data-driven learning of fundamental diagrams for a roadway segment, we show how exogenous factors beyond the lane configuration of a roadway can be detected as impacting the speed-flow characteristics of the roadway segment. In this experiment, we simulated two extreme cases of a single-travel lane roadway segment in a city network with curbside parking. In Scenario 1, all drivers are fully compliant with parking regulations, whereas in Scenario 2, a percentage of drivers will double-park and block traffic for brief periods of time, simulating the impact of passenger pick-up/drop-off activities on the long-term speed-flow characteristics of the roadway. As such, Scenario 2 exhibits a greater degree of congestion and state variance than that of Scenario 1. 

\subsection{One-variable FDs}
Figure \ref{fig:single_variable} shows raw speed-density and flux-density data pairs with three candidate FDs overlaid for Scenario 1 (left column) and Scenario 2 (right column). The three overlaid FDs correspond to (i) a baseline least-squares regression of Greenshield's FD (Eq. \ref{eq:green}), (ii) the neural ODE formulation of Greenshield's FD (Eq. \ref{eq:green_node}), and (iii) the black-box Neural ODE (Eq. \ref{eq:single}). The difference in objectives between the baseline least-squares regression and the trajectory reconstruction problems becomes apparent in examining the free-flow speeds and jam densities of the three FDs. The baseline FD has the fastest free-flow speed, largest carrying capacity, and greatest critical density of the three FDs for both scenarios. However, as evidenced by the reconstruction error metrics in Table \ref{tab:metrics}, the baseline FD is the least performant in reconstructing the trajectories of the training and test data across both data splits and scenarios. Furthermore, the least-squares regression predicts an increase in carrying capacity from Scenario 1 to Scenario 2. Contrasting the baseline FD, recasting the same model prior (i.e. the functional form of Eq. \ref{eq:green}) as a trajectory reconstruction problem results in a FD that out performs than the baseline FD in both scenarios. The improvement over the baseline is small for Scenario 1. The improvement over baseline for Scenario 2 is more significant, with an approximate order-of-magnitude improvement for the test data. In eliminating the Greenshield FD prior, we are left with the black-box Neural ODE model, albeit still kinematically consistent. This formulation has the greatest degree of expressivity; this is apparent in its learned shape. The flexibility of this FD also gives the best trajectory reconstruction performance of the investigated one-variable FDs.

\begin{figure*}[] 
        \centering
            \begin{overpic}[width=6in]{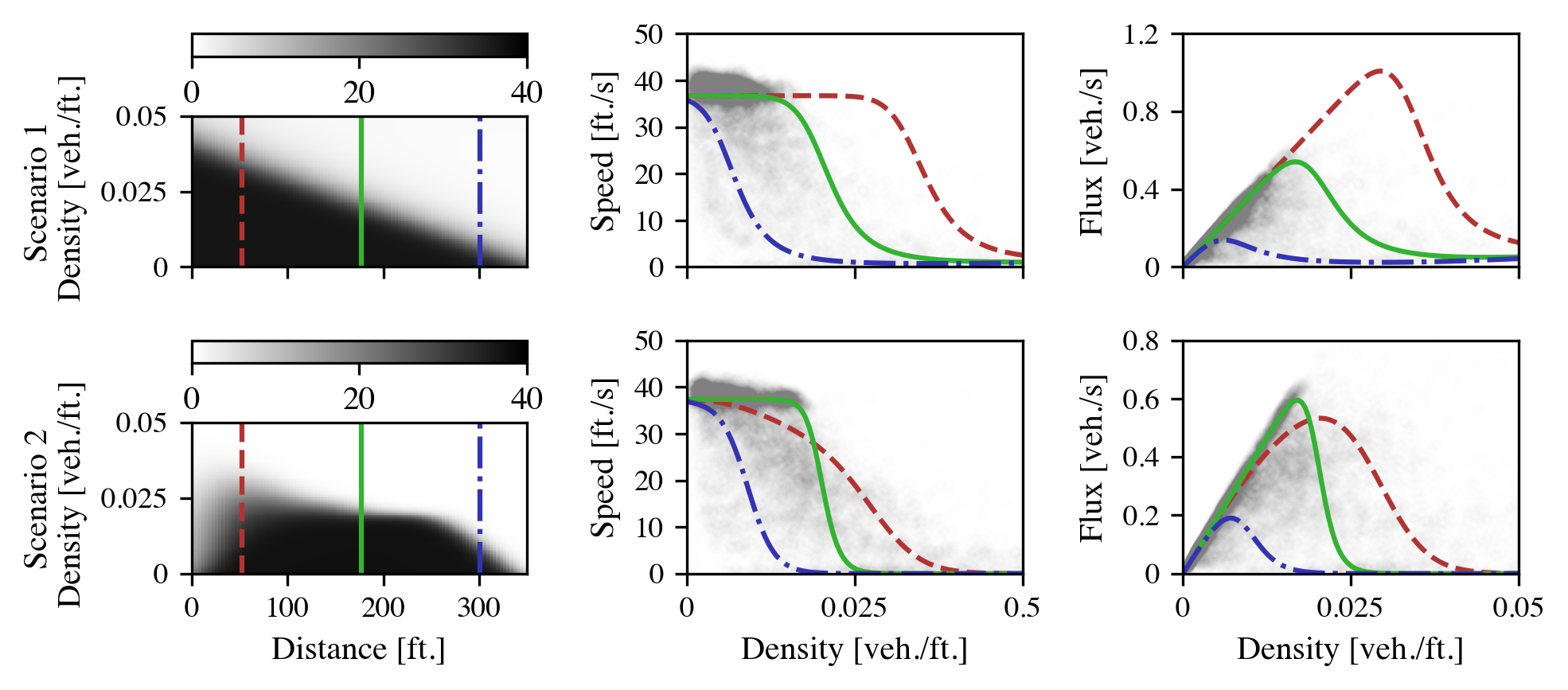} 
            \put(8,39.5){(a)}
            \put(62,39.5){(b)}
            \put(94,39.5){(c)}
            \put(8,20){(d)}
            \put(62,20){(e)}
            \put(94,20){(f)}

	    \end{overpic}  
	    \caption{Learned parameterized neural network-based fundamental diagrams for Scenarios 1 and 2. In this learning task, the networks were parameterized by the location along the length of the roadway; i.e. $\NN(\rho,x)$. The left column is a heatmap of the learned speed-density-distance relationship. The learned 2-dimensional fundamental diagrams (left) are sliced at 56ft, 175ft, and 300ft; these speed-density profiles at those spatial locations are shown in the middle column. The corresponding flux-density relationships are displayed in the right column.}
		\label{fig:param}
\end{figure*}
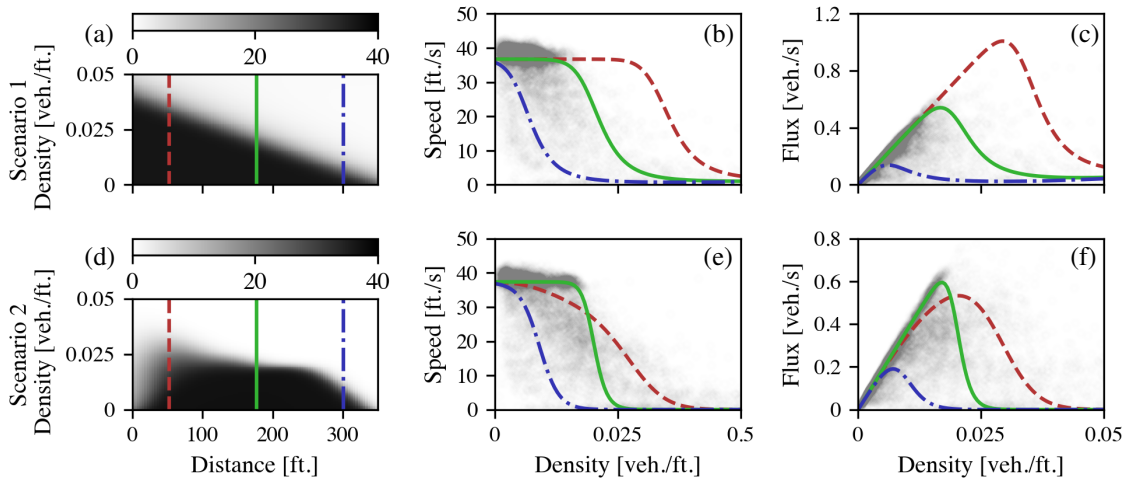

\subsection{Parameterized FD} \label{sec:param}
A natural extension of the single-variable black-box neural ODE is the parameterization of the network by location ($x$) as in Eq. \ref{eq:param}. The trained 2-dimensional FDs for Scenarios 1 and 2 are shown in Figure \ref{fig:param}. A representative FD at a location along the roadway can be visualized by fixing the spatial input to the network; e.g. $u_0 \cdot \NN\left(\rho/\rho_j,x=50 ft.\right)$. The middle column of Fig. \ref{fig:param} shows three such speed-density profiles at locations $x=56$ ft., $x=175$ ft., and $x=300$ ft. for both Scenarios. Similarly, the corresponding flux-density profiles are given in the right column. For Scenario 1, the learned FD has a generally linear trend with increasing distance: as a vehicle approaches the end of the roadway (an intersection with a stoplight for these scenarios), the \textit{local} carrying capacity is reduced, as is the \textit{local} critical density. This appears as a lateral shift of the FD to the left with increasing distance, as depicted in Fig. \ref{fig:param}b. For that of Scenario 2, the profiles show a more intricate FD: the middle section of the roadway ($100<x<250 $ ft.) is relatively insensitive to the spatial location and possesses the highest carrying capacity of anywhere along the roadway. However, the beginning and end of the roadway show reduced capacities. The behavior at the end of the roadway is consistent with that of Scenario 1; that is, further reduced capacity as one approaches the intersection and stoplight. 

These parameterized FDs are the most performant models of those examined in this work. The reconstruction error for Scenario 1 is about an order of magnitude better than the baseline FD. For Scenario 2, the improvement in the error metric is roughly two orders of magnitude.

\section{Discussion} \label{sec:discussion}

In this work, we have recast the Fundamental Diagram curve-fitting task as one of probe vehicle trajectory reconstruction through parameterized Neural Ordinary Differential Equations. While the presented methodology has many benefits, we emphasize three in particular: (i) the connection between macroscopic and microscopic FDs, (ii) the introduction of \textit{contextual} FDs, and (iii) trajectory reconstruction as opposed to standard least-squares regression. 

\subsection{Fundamental Diagram Granularity}
A \textit{microscopic} diagram is one whose spatial resolution is finer than the minimal link (i.e. a city block street bounded by intersections) of a roadway. A \textit{macroscopic} diagram is one whose shape represents the averaged or otherwise aggregated behavior of vehicles on a roadway link. The model specified by Eq. \ref{eq:single} learns a \textit{macroscopic} fundamental diagram. Similarly, the model specified by Eq. \ref{eq:param} learns a continuum of \textit{microscopic} fundamental diagrams. The macro/micro distinction is by construction for these models; the single-variable Neural ODE-based FDs lump all effects into the learned or tuned (in the case of Greenshield's FD or other well-established FDs) models. The parameterized FD instead explicitly encodes space - as such, the FD can be evaluated at any point along the length of the roadway. Microscopic fundamental diagrams are useful for highly local control decisions, such as adaptive cruise control in connected environments, where the speed-flow properties of a roadway along a local segment can be leveraged to improve total traffic flow, and applied in parallel from segment to segment.

\subsection{Parametric Fundamental Diagrams}
A key contribution of this work is a universally applicable methodology for parameterizing fundamental diagrams. Both macroscopic and microscopic modeling regimes can be easily extended to include other exogenous inputs. These inputs could include roadway characteristics like parking demand, inclement weather, traffic light scheduling, etc. simply by adding these as additional inputs to the respective neural networks.

\subsection{Trajectory Reconstruction}
The trajectory reconstruction problem features prominently an objective that departs from conventional least-squares curve-fitting tasks often used for FD estimation and/or calibration. As demonstrated by comparing the baseline least-squares regression of Greenshield's FD to Scenario 1 and 2 data, the least-squares regression does not reproduce vehicle trajectories well. Note, however, that the baseline regression was performed naively; i.e. without preferentially sampling data points.

Several modeling choices were implicitly made in this optimization task. First, for a collection of vehicle trajectories in a dataset, not each vehicle experiences the same dwell time in the roadway. For example, in these simulation studies, most vehicles either traversed the length of the link in roughly 8 seconds or were stopped for some period of time (e.g. experiencing a parking action or a red traffic signal), increasing dwell time to about 20 to 30 seconds. For this work, we weighted each vehicle trajectory equally in its contribution to the overall objective. An alternative would be to weight each individual data point equally, regardless of the particular vehicle trajectory. Additionally, the associated IVP for each trajectory has a time span equal to the dwell time. For the trajectory minimization objective, the divergence of the reference trajectory and model trajectory is intimately tied to preceding time steps. Training in this manner therefore has an implicit bias of fitting data points closer to the beginning of the time series as opposed to later ones. If desired, this bias can be eliminated by decomposing the vehicle trajectories into their minimum temporal resolution ($\Delta t$) and re-running the training process outlined in Section \ref{sec:training} with the added 1-step vehicle trajectories.

For this study, we used both static sensor data (Eulerian reference frame) as well as probe vehicle data (Lagrangian reference frame). To successfully perform the trajectory reconstruction problem, speed-density data pairs are needed along the path of the vehicle. While vehicle data provides a time series for speed along this path, density is unknown. The static sensor array (Fig. \ref{fig:simulator}) provides an estimate for flux at each one of the sensor locations, thus giving an estimate for density when combined with vehicle velocity. In general, the presented methodology is agnostic as to the data source - all that is required are co-located density-velocity data points along the vehicle path. For example, should a probe vehicle be equipped with sensors for measuring following distance, a car-following FD can be constructed using the same methodology presented herein. 

Lastly, we chose to constrain the scope of this work to exclusively first-order FDs; that is, disregarding vehicle acceleration dynamics. The Neural ODE framework is flexible: to include dynamics to arbitrary order (for example, to recast the 2-variable parameterized FD of Eq. \ref{eq:param} as acceleration-dependent) we can define the first-order system:

\begin{align}
    \frac{dx}{dt} &= v \nonumber \\
    \frac{d v }{dt} &= \NN \left( x, v, \frac{\rho}{\rho_j}; \theta \right) .
\end{align}

\section{Conclusion}
This work presents \textit{contextual} Fundamental Diagrams (FD); those that are parameterized by exogenous inputs. These FDs are constructed through Neural Ordinary Differential Equations (Neural ODEs), where we recast the typical curve-fitting task as a trajectory reconstruction task. We find that these parametric FDs are generally more expressive and performant in reproducing observed behavior of a collection of probe vehicles. This methodology allows for constructing diagrams of arbitrary granularity; from macroscopic diagrams based on mean flow properties of a link to microscopic diagrams that continuously evolve with distance travelled down a link. We envision the presented ideas will have broad applicability in control settings, particularly those involving Connected Autonomous Vehicles, where highly local control decisions are necessary to improve total traffic flow and can be applied in parallel from segment to segment.

\section*{Acknowledgments}

Pacific Northwest National Laboratory is operated by Battelle Memorial Institute for the U.S. Department of Energy under Contract No. DE-AC05-76RL01830. This work was supported by the U.S. Department of Energy Vehicle Technologies Office.

\bibliographystyle{plain} 
\bibliography{traffic}

\end{document}